\documentclass{article}

\usepackage{arxiv}

\usepackage[utf8]{inputenc} 
\usepackage[T1]{fontenc}    
\usepackage{hyperref}       
\usepackage{url}            
\usepackage{booktabs}       
\usepackage{amsfonts}       
\usepackage{nicefrac}       
\usepackage{microtype}      
\usepackage{lipsum}
\usepackage{graphicx}
\graphicspath{ {./images/} }

\title{Code-Mixed Telugu-English Hate Speech Detection}

\author{
 Santhosh Kakarla \\
  George Mason University\\
  \texttt{skakarl3@gmu.edu} \\
   \And
 Gautama Shastry Bulusu Venkata \\
  George Mason University\\
  \texttt{sbulusuv@gmu.edu} \\
}

\begin{document}
\maketitle
\begin{abstract}
Hate speech detection in low-resource languages like Telugu is a growing challenge in NLP. This study investigates transformer-based models, including TeluguHateBERT, HateBERT, DeBERTa, Muril, IndicBERT, Roberta, and Hindi-Abusive-MuRIL, for classifying hate speech in Telugu. We fine-tune these models using Low-Rank Adaptation (LoRA) to optimize efficiency and performance. Additionally, we explore a multilingual approach by translating Telugu text into English using Google Translate to assess its impact on classification accuracy.

Our experiments reveal that most models show improved performance after translation, with DeBERTa and Hindi-Abusive-MuRIL achieving higher accuracy and F1 scores compared to training directly on Telugu text. Notably, Hindi-Abusive-MuRIL outperforms all other models in both the original Telugu dataset and the translated dataset, demonstrating its robustness across different linguistic settings. This suggests that translation enables models to leverage richer linguistic features available in English, leading to improved classification performance. The results indicate that multilingual processing can be an effective approach for hate speech detection in low-resource languages. These findings demonstrate that transformer models, when fine-tuned appropriately, can significantly improve hate speech detection in Telugu, paving the way for more robust multilingual NLP applications.
\end{abstract}


\section{Introduction}
Telugu, spoken by over 80 million people in India, presents unique linguistic and syntactic challenges that make hate speech detection complex. Traditional approaches, such as rule-based systems and classical machine learning models, fail to capture the intricate contextual dependencies of hate speech. Recent advancements in NLP, particularly transformer-based models, offer promising solutions for tackling this issue. This study explores various transformer architectures and compares their performance in classifying hate speech in Telugu. Additionally, we investigate the impact of translating Telugu hate speech into English, demonstrating that translation can enhance classification accuracy by enabling models to utilize richer linguistic features available in English. By leveraging transformer models and multilingual approaches, this research aims to bridge the gap in hate speech detection for low-resource languages and contribute to the development of more inclusive and effective NLP applications.

\section{Related Work}
Hate speech detection has been extensively studied in NLP, primarily for high-resource languages such as English. Early methods relied on lexicon-based approaches and classical machine learning models like Naïve Bayes and Support Vector Machines (SVMs). These models, while effective for basic classification, struggled with contextual nuances. Deep learning techniques, particularly convolutional neural networks (CNNs) and recurrent neural networks (RNNs), improved performance by capturing sequential dependencies in text. However, the advent of transformer-based architectures, such as BERT, HateBERT, and XLM-Roberta, has significantly advanced the field by leveraging contextual embeddings and attention mechanisms. Recent works have demonstrated the effectiveness of fine-tuning pre-trained transformers for hate speech detection across multiple languages. Additionally, multilingual models such as MURIL and IndicBERT have shown promise for low-resource Indian languages. Our study builds upon these prior works by focusing on hate speech detection in Telugu using multilingual transformer models. Furthermore, we explore the impact of translation-based classification and compare it with direct training on Telugu text. Our approach integrates LoRA fine-tuning and model ensembling to optimize classification accuracy while addressing the challenges associated with linguistic variations in Telugu hate speech.

\section{Methodology}
\subsection{Dataset Preparation}

We utilize a dataset containing Telugu text labeled as "hate" or "non-hate." The dataset undergoes preprocessing steps, including tokenization, padding, and truncation to fit transformer model requirements. The training set consists of a nearly balanced distribution of hate and non-hate speech instances, as illustrated in the figure below. Similarly, the test set maintains an even split between the two categories, ensuring a fair evaluation of model performance.

\begin{figure}[ht]
    \includegraphics[width=0.8\textwidth]{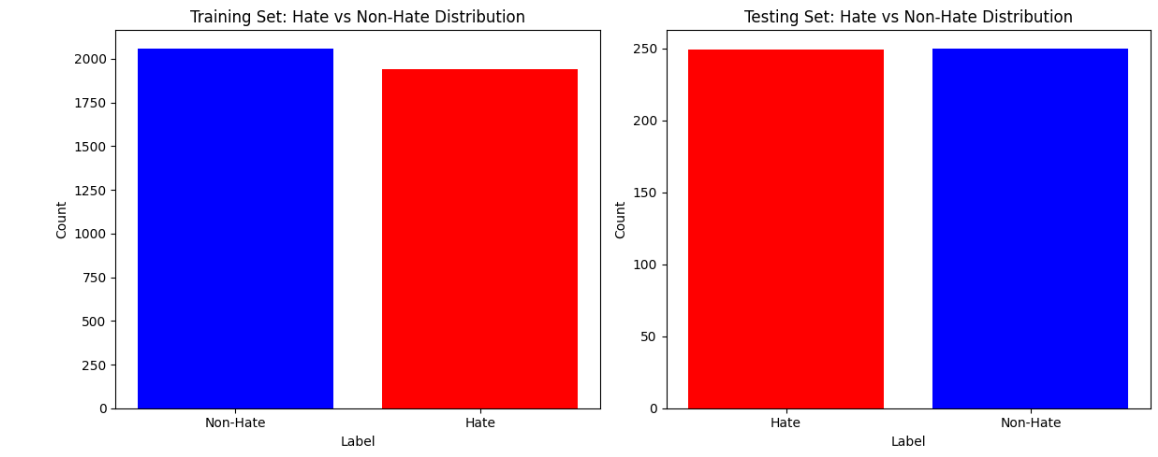}
    \caption{Distribution of hate and non-hate speech in the training and testing datasets. The dataset is balanced to ensure fair evaluation of model performance.}
\end{figure}

\subsection{Model Selection and Fine-tuning with LoRA}

To effectively classify hate speech in Telugu, we employ a range of transformer-based models known for their proficiency in text classification. The models used in this study include TeluguHateBERT, a fine-tuned BERT model specifically trained on Telugu text, HateBERT, which has been pre-trained for hate speech detection, and DeBERTa, a transformer model with an enhanced attention mechanism that provides better contextual understanding. Additionally, we explore multilingual models such as Muril and IndicBERT, which are optimized for Indian languages, Roberta, a robust transformer-based model that demonstrated superior performance among individual models, and Hindi-Abusive-MuRIL, which performed exceptionally well across both the original Telugu dataset and the translated dataset, demonstrating strong generalization ability.

Fine-tuning these models effectively requires optimizing their parameters while maintaining computational efficiency. To achieve this, we utilize Low-Rank Adaptation (LoRA), which reduces the number of trainable parameters, making fine-tuning more efficient without compromising model performance. Training is conducted using the AdamW optimizer with a learning rate of 2e-5, a batch size of 32, and 25 epochs. LoRA enables the transformer models to learn specialized patterns in Telugu hate speech data while maintaining their pre-trained general language understanding. This approach significantly enhances the model’s ability to detect nuanced expressions of hate speech in Telugu while keeping computational demands manageable.
Low-Rank Adaptation (LoRA) is implemented to reduce the number of trainable parameters while maintaining model performance.

\subsection{Telugu-to-English Translation}

To explore the impact of multilingual processing on hate speech detection, we translate the Telugu dataset into English using Google Translate. This approach aims to leverage pre-trained English-based hate speech models and assess whether translation improves classification performance. The translated dataset undergoes the same preprocessing steps as the original Telugu dataset, including tokenization, padding, and truncation.

Once translated, we fine-tune the same transformer-based models (Roberta, TeluguHateBERT, HateBERT, Hindi-Abusive MuRIL, DeBERTa, etc) on the English dataset and evaluate their performance. The hypothesis is that translating Telugu hate speech into English could improve classification accuracy by utilizing more resource-rich English-language models. Our findings indicate that translation generally improves accuracy but leads to a decrease in F1-score for several models. The decline in F1-score can be attributed to the loss of linguistic and contextual nuances during translation, as Telugu-specific hate expressions may not have direct English equivalents. These results highlight the importance of refining translation techniques and using hybrid approaches for more effective hate speech detection in low-resource languages.

\section{Results}

The evaluation of the models provides insights into their effectiveness in detecting hate speech in Telugu. The primary metrics used for analysis are accuracy and F1-score (both macro and weighted), which measure the overall correctness and balance between precision and recall. The results indicate notable variations in performance among the multilingual models.

\begin{table}[h]
    \centering
    \resizebox{0.8\textwidth}{!}{%
    \begin{tabular}{|l|c|c|c|}
        \hline
        \textbf{Model} & \textbf{Test Accuracy} & \textbf{Test F1 Weighted} & \textbf{Test F1 Macro} \\
        \hline
        Roberta         & 0.6773  & 0.6772  & 0.6772  \\
        DeBERTa         & 0.4268  & 0.4148  & 0.4150  \\
        Muril           & 0.4228  & 0.4114  & 0.4115  \\
        TeluguHateBERT  & 0.4388  & 0.4250  & 0.4251  \\
        IndicBERT       & 0.3687  & 0.3632  & 0.3633  \\
        HateBERT        & 0.4509  & 0.4412  & 0.4414  \\
        Hindi-Abusive-MuRIL & 0.7314 & 0.7314 & 0.7314 \\
        \hline
    \end{tabular}} 
    \caption{Performance of Transformer Models on Translated Telugu Hate Speech Dataset}
\end{table}

The results indicate that Hindi-Abusive-MuRIL achieved the highest accuracy and F1-score among all models, demonstrating its ability to generalize well for Telugu hate speech detection. Roberta followed as the second-best performer, benefiting from its robust pretraining. TeluguHateBERT and HateBERT exhibited moderate performance, while IndicBERT and Muril had the lowest accuracy and F1-scores. The weaker performance of multilingual models suggests that they may not be as well-optimized for Telugu-specific nuances.

The relatively lower performance of DeBERTa, despite being a powerful transformer model, suggests that it may require additional domain-specific fine-tuning for Telugu hate speech detection. The results reinforce that models specifically trained on abusive speech (such as Hindi-Abusive-MuRIL) tend to generalize better even for other Indian languages, making them an effective option for hate speech detection tasks.

When the Telugu dataset was translated into English, accuracy improved for most models, but F1-scores generally declined. Hindi-Abusive-MuRIL and DeBERTa showed the highest accuracy improvements post-translation, suggesting that English-based models can leverage richer linguistic patterns from well-trained English embeddings. However, despite these improvements in accuracy, the drop in F1-scores highlights the challenges introduced by translation. The decline in F1-score after translation is primarily due to loss of linguistic and contextual nuances. Certain Telugu-specific hate expressions and slang terms may not have direct English equivalents, leading to misclassification of hate speech as non-hate speech. Additionally, translation may introduce neutralized phrasing, which can affect recall and precision, thereby lowering the F1-score.

\begin{table}[h]
    \centering
    \resizebox{0.8\textwidth}{!}{%
    \begin{tabular}{|l|c|c|c|}
        \hline
        \textbf{Model} & \textbf{Test Accuracy} & \textbf{Test F1 Weighted} & \textbf{Test F1 Macro} \\
        \hline
        Roberta         & 0.5050  & 0.3823  & 0.3829  \\
        DeBERTa         & 0.6032  & 0.5657  & 0.5654  \\
        Muril           & 0.5010  & 0.3344  & 0.3338  \\
        TeluguHateBERT  & 0.5010  & 0.3344  & 0.3338  \\
        IndicBERT       & 0.5010  & 0.3344  & 0.3338  \\
        HateBERT        & 0.4669  & 0.4665  & 0.4665  \\
        Hindi-Abusive-MuRIL & 0.6993 & 0.6990 & 0.6990 \\
        \hline
    \end{tabular}}
    \caption{Performance of Transformer Models on Translated Telugu Hate Speech Dataset}
\end{table}

These findings confirm that model performance varies depending on training conditions, with Hindi-Abusive-MuRIL demonstrating strong generalization in both original and translated datasets. Future research should focus on refining translation techniques, expanding Telugu-specific datasets, and developing better pretraining strategies to enhance performance further.

\section{Conclusion}

This study evaluated transformer-based models for hate speech detection in Telugu, with and without translation to English. The results demonstrate that Hindi-Abusive-MuRIL consistently outperformed other models in both Telugu and translated datasets, making it the most robust option for hate speech detection. Translation-based classification improved accuracy for most models but led to lower F1-scores, indicating that while translation helps models leverage English-based embeddings, it also results in the loss of linguistic nuances crucial for identifying hate speech. These findings underscore the importance of language-specific fine-tuning and hybrid approaches that combine multilingual and monolingual training strategies. Future work should focus on expanding the Telugu hate speech dataset, optimizing translation techniques to preserve context, and developing cross-lingual embeddings that retain language-specific characteristics while enhancing generalization across multiple languages.

\section{Limitations}

Despite the promising results, there are several limitations to this study. First, the dataset size is relatively small compared to large-scale hate speech datasets available for high-resource languages like English. A larger dataset with diverse hate speech examples could enhance model generalization. Second, while translation-based classification was explored, it relied on Google Translate, which may not accurately capture the nuances of Telugu hate speech. Finally, computational constraints limited our ability to experiment with larger models and alternative fine-tuning techniques, which could further improve classification accuracy.

\bibliographystyle{unsrt}  


\end{document}